\documentclass{article}

\usepackage{PRIMEarxiv}

\usepackage[utf8]{inputenc} 
\usepackage[T1]{fontenc}    
\usepackage{hyperref}       
\hypersetup{
    colorlinks=true,
    linkcolor=blue,
    filecolor=magenta,
    urlcolor=cyan,
    pdftitle={BioNeMo Framework},
    pdfpagemode=FullScreen,
}
\usepackage{url}            
\usepackage{booktabs}       
\usepackage{amsfonts}       
\usepackage{nicefrac}       
\usepackage{microtype}      
\microtypesetup{protrusion=false}
\usepackage{lipsum}
\usepackage{xcolor}
\usepackage{fancyhdr}       
\usepackage{graphicx}       
\usepackage{setspace}       
\usepackage[backend=biber]{biblatex}
\graphicspath{{media/}}     

\title{BioNeMo Framework: A Modular, High-performance Library for AI Model Development in Drug Discovery}

\author{
\begin{tabular}{c}
\textbf{Peter St. John}$^{a}$, \textbf{Dejun Lin}$^{a}$, \textbf{Polina Binder}$^{a}$, \textbf{Malcolm Greaves}$^{a}$, \textbf{Vega Shah}$^{a}$, \textbf{John St. John}$^{a}$, \\
\textbf{Adrian Lange}$^{b}$, \textbf{Patrick Hsu}$^{c}$, \textbf{Rajesh Illango}$^{c}$, \textbf{Arvind Ramanathan}$^{d}$, \textbf{Anima Anandkumar}$^{e}$, \\
\textbf{David H Brookes}$^{f}$, \textbf{Akosua Busia}$^{f}$, \textbf{Abhishaike Mahajan}$^{f}$, \textbf{Stephen Malina}$^{f}$, \textbf{Neha Prasad}$^{f}$, \\
\textbf{Sam Sinai}$^{f}$, \textbf{Lindsay Edwards}$^{g}$, \textbf{Thomas Gaudelet}$^{g}$, \textbf{Cristian Regep}$^{g}$, \textbf{Martin Steinegger}$^{h}$, \\
\textbf{Burkhard Rost}$^{i}$, \textbf{Alexander Brace}$^{jk}$, \textbf{Kyle Hippe}$^{jk}$, \textbf{Luca Naef}$^{l}$, \textbf{Keisuke Kamata}$^{m}$, \\
\textbf{George Armstrong}$^{a}$, \textbf{Kevin Boyd}$^{a}$, \textbf{Zhonglin Cao}$^{a}$, \textbf{Han-Yi Chou}$^{a}$, \textbf{Simon Chu}$^{a}$, \\
\textbf{Allan dos Santos Costa}$^{a}$, \textbf{Sajad Darabi}$^{a}$, \textbf{Eric Dawson}$^{a}$, \textbf{Kieran Didi}$^{a}$, \textbf{Cong Fu}$^{a}$, \\
\textbf{Mario Geiger}$^{a}$, \textbf{Michelle Gill}$^{a}$, \textbf{Darren J Hsu}$^{a}$, \textbf{Gagan Kaushik}$^{a}$, \textbf{Maria Korshunova}$^{a}$, \\
\textbf{Steven Kothen-Hill}$^{a}$, \textbf{Youhan Lee}$^{a}$, \textbf{Meng Liu}$^{a}$, \textbf{Micha Livne}$^{a}$, \textbf{Zachary McClure}$^{a}$, \\
\textbf{Jonathan Mitchell}$^{a}$, \textbf{Alireza Moradzadeh}$^{a}$, \textbf{Ohad Mosafi}$^{a}$, \textbf{Youssef Nashed}$^{a}$, \textbf{Saee Paliwal}$^{a}$, \\
\textbf{Yuxing Peng}$^{a}$, \textbf{Sara Rabhi}$^{a}$, \textbf{Farhad Ramezanghorbani}$^{a}$, \textbf{Danny Reidenbach}$^{a}$, \textbf{Camir Ricketts}$^{a}$, \\
\textbf{Brian C Roland}$^{a}$, \textbf{Kushal Shah}$^{a}$, \textbf{Tyler Shimko}$^{a}$, \textbf{Hassan Sirelkhatim}$^{a}$, \textbf{Savitha Srinivasan}$^{an}$, \\
\textbf{Abraham C Stern}$^{a}$, \textbf{Dorota Toczydlowska}$^{a}$, \textbf{Srimukh Prasad Veccham}$^{a}$, \textbf{Niccolò Alberto Elia Venanzi}$^{a}$, \\
\textbf{Anton Vorontsov}$^{a}$, \textbf{Jared Wilber}$^{a}$, \textbf{Isabel Wilkinson}$^{a}$, \textbf{Wei Jing Wong}$^{a}$, \textbf{Eva Xue}$^{a}$, \\
\textbf{Cory Ye}$^{a}$, \textbf{Xin Yu}$^{a}$, \textbf{Yang Zhang}$^{a}$, \textbf{Guoqing Zhou}$^{a}$, \textbf{Becca Zandstein}$^{a}$, \\
\textbf{Alejandro Chacon}$^{a}$, \textbf{Prashant Sohani}$^{a}$, \textbf{Maximilian Stadler}$^{a}$, \textbf{Christian Hundt}$^{a}$, \textbf{Feiwen Zhu}$^{a}$, \\
\textbf{Christian Dallago}$^{ao}$, \textbf{Bruno Trentini}$^{ap}$, \textbf{Emine Kucukbenli}$^{a}$, \textbf{Saee Paliwal}$^{a}$, \textbf{Timur Rvachov}$^{a}$, \textbf{Eddie Calleja}$^{a}$, \\
\textbf{Johnny Israeli}$^{a}$, \textbf{Harry Clifford}$^{a}$, \textbf{Risto Haukioja}$^{a}$, \textbf{Nicholas Haemel}$^{a}$, \\ \textbf{Kyle Tretina}$^{a*}$, 
\textbf{Neha Tadimeti}$^{a*}$, \textbf{Anthony B Costa}$^{a*}$ \\
\end{tabular}
}
\begin{document}
\maketitle
\vspace{-2em} 
\begin{abstract}
Artificial Intelligence models encoding biology and chemistry are opening new routes to high-throughput and high-quality in-silico drug development. However, their training increasingly relies on computational scale, with recent protein language models (pLM) training on hundreds of graphical processing units (GPUs). We introduce the BioNeMo Framework to facilitate the training of computational biology and chemistry AI models across hundreds of GPUs. Its modular design allows the integration of individual components, such as data loaders, into existing workflows and is open to community contributions. We detail technical features of the BioNeMo Framework through use cases such as pLM pre-training and fine-tuning. On 256 NVIDIA A100s, BioNeMo Framework trains a three billion parameter BERT-based pLM on over one trillion tokens in 4.2 days. The BioNeMo Framework is open-source and free for everyone to use.
\end{abstract}

\begingroup
  \renewcommand{\thefootnote}{}
  \footnotetext{\textbf{Affiliations:} 
        $^a$NVIDIA, 
        $^b$A-Alpha Bio Inc., 
        $^c$Arc Institute, 
        $^d$Argonne National Laboratory, 
        $^e$Caltech, 
        $^f$Dyno Therapeutics, 
        $^g$Relation Therapeutics, 
        $^h$Seoul National University, 
        $^i$Technical University of Munich, 
        $^j$University of Chicago, 
        $^k$Argonne National Laboratory, 
        $^l$VantAI, 
        $^m$Weights \& Biases, 
        $^n$Stanford University, 
        $^o$Duke University,
        $^p$University of Oxford.
        $^*$\textbf{Correspondence to}: {Anthony B Costa: acosta@nvidia.com; Neha Tadimeti: ntadimeti@nvidia.com;  Kyle Tretina: ktretina@nvidia.com.}}
  \addtocounter{footnote}{-1}
\endgroup

\setlength{\leftskip}{2em} 

\keywords{artificial intelligence \and drug discovery \and deep learning \and framework \and biomolecules}

\section{Introduction}
The field of Artificial Intelligence (AI) has revolutionized computational biology and chemistry, culminating in the 2024 Nobel Prize in Chemistry for advancements in computational protein design and protein structure prediction. These breakthroughs have been driven by the rapid growth of high-throughput experimental biology \cite{behjati2013next}, extensive sequence databases \cite{uniprot2023uniprot,lee2024petabase}, and significant progress in large-scale neural network architectures \cite{vaswani2017attention} leading to Large Language Models (LLMs). As such, protein Language Models (pLMs) \cite{elnaggar2021prottrans, Lin2023-sb}, trained on vast protein sequence databases \cite{uniprot2023uniprot}, are now achieving unprecedented accuracy in tasks such as structure prediction \cite{Lin2023-sb}, function prediction \cite{bernhofer2021predictprotein}, and de novo protein design \cite{ferruz2023sequence}.

With recent understanding of LLM scaling laws, pLMs have been growing in size to improve performance, with ESM3 featuring 98 billion parameters and requiring 10\textsuperscript{24} floating-point operations (FLOPS) to train \cite{Lin2023-sb, Kaplan2020-en}. While improvements in training data quality may also help improve performance \cite{Fournier2024-fp}, we believe that frontier pLMs will continue to increase in size and training requirements. Beyond protein design, LLMs are also driving new techniques in chemical and material discovery \cite{Boiko2023-mj, Kononova2021-cq}. 

AI-driven molecular research is driving a rapid increase in the need for specialized machine-learning frameworks tailored to drug discovery. While traditional tools like PyTorch\cite{Paszke2019-ch}, TensorFlow\cite{tensorflow2015-whitepaper}, and JAX\cite{jax2018github} have been successful in advancing deep learning, these frameworks are not optimized for the complex biomolecular data and large-scale models required in drug discovery. Many of these frameworks are excellent for general-purpose deep learning but often require intricate adaptation for tasks such as protein structure prediction or chemical reaction modeling, resulting in trade-offs between flexibility and performance. Efficiently training very large models requires careful tuning of data and model-parallelism strategies and requires careful use of fused CUDA kernels to minimize memory usage and maximize training throughput \cite{Kuchaiev2019-am}\cite{Narayanan2021-qc}. High-performance data loaders are also a critical component of a pre-training pipeline, as ensuring that GPUs are fully utilized requires that pre-processing and memory access is done as efficiently as possible. Minimizing energy consumption and maximizing model performance are especially important as model training increases in scale. Further progress in these fields also requires that these techniques remain accessible to smaller research groups and domain scientists in chemical and biological research areas.

In this paper, we present the BioNeMo Framework, a suite of open-source python packages designed to simplify the process of training high-performance AI models in the biomolecular and chemical space. These libraries build on top of NVIDIA NeMo\cite{Kuchaiev2019-am} and Megatron-LM \cite{Narayanan2021-qc} to optimize training and inference throughput, and include high-performance data loaders for both protein sequence sampling and single-cell data. In this release, BioNeMo Framework includes the ESM-2 \cite{Lin2023-sb} and Geneformer \cite{Theodoris2023-np} models, where we demonstrate an over 2x improvement in training throughput over PyTorch implementations and near-linear scaling in multi-device training up to 256 GPUs. We also discuss future directions for the BioNeMo Framework and how external contributors can extend the framework to accelerate additional families of models.

\section{Overview of BioNeMo Framework}
\label{sec:headings}
NVIDIA BioNeMo Framework is a collection of programming tools, libraries, and models for computational drug discovery. It accelerates the most time-consuming and costly stages of building and adapting biomolecular AI by providing domain-specific, optimized models and tooling that can be easily integrated into any GPU-based compute environment.

The BioNeMo Framework codebase is designed with a sub-project structure that facilitates ease of contribution and focus among its users. While the entire codebase can be installed as a cohesive unit, it is also segmented into independently installable components, referred to as sub-projects, organized by their specific purposes and scopes. This modular architecture enhances the capability for extensibility, allowing new contributions to take two forms: the creation of new sub-projects (wide contributions) or the extension of existing sub-projects (deep contributions).

At the center of BioNeMo lies bionemo-core, which provides essential interfaces and common data processing and model building blocks built around PyTorch and the Lightning framework. This core dependency supports all other sub-projects within the framework. For instance, the bionemo-llm sub-package is built upon NVIDIA’s NeMo and Megatron libraries, providing bio-specific customizations and foundational model architectures. The bionemo-geometric sub-package is developing support for Graph Neural Networks(GNNs)  leveraging PyTorch Geometric, highlighting the breadth of innovations possible within the BioNeMo ecosystem.

Specific models and data loaders are encapsulated within their sub-packages, enabling users to selectively incorporate only the components relevant to their particular use cases. This modular design not only streamlines customization but also enhances usability.

To foster easy integration into private projects, BioNeMo sub-projects are designed to be hosted on the Python Package Index\footnote{More details available at https://pypi.org/} (PyPI). Docker images enable rapid development, allowing users to set up a complete environment quickly. Internal tools within the repository further assist users in creating and integrating their own Python projects into workflows, streamlining development.

BioNeMo Framework currently supports building large-scale BERT-based models through NVIDIA NeMo Megatron, a scalable framework for developing custom LLMs, multimodal, and speech AI. BioNeMo Framework uses NeMo Megatron to build biomolecular BERT model support.

\section{Functionalities enabled through BioNeMo Framework}

\subsection{BERT Model Architecture Support}
Two example BERT implementations -- ESM-2 for protein sequence modeling and Geneformer for single-cell expression modeling -- are currently available in the BioNeMo Framework codebase. The ESM-2 model is packaged into the submodule \textit{bionemo-esm2}, and the Geneformer model is packaged into an independent submodule \textit{bionemo-geneformer}.

Users can install and use either of these packages off-the-shelf and perform training, fine-tuning, and inference with custom data. Each sub-package has its own independent \textit{src/} directory, which hosts data modules, data tokenizers, and model implementations. Customizing the models can be as simple as making a copy of the sub-package and swapping out the modules, such as data loaders, with custom implementations. 

Users can also easily customize loss functions, change different layers in the model architecture, etc., using the reference packages as examples. 

One of BioNeMo Framework's most significant advantages, beyond its ease of use and customizability, is its seamless, ready-to-use scalability. With built-in support for data, model, tensor, and pipeline parallelism, scaling models across devices or increasing the number of parameters is simplified. Users only need to adjust the appropriate configuration parameters, requiring minimal or no changes to their code.

\subsection{Benchmarks and Results}
\begin{figure}
    \centering
    \includegraphics[width=0.5\linewidth]{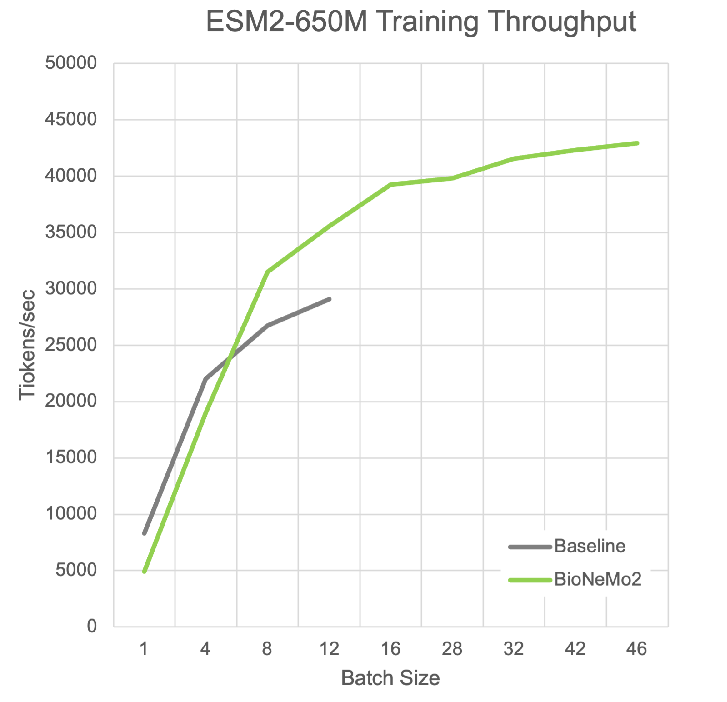}
    \caption{The BioNeMo Framework improves single-node memory usage and throughput compared to a standard PyTorch implementation. The ESM-2 650M parameter model trained with Hugging Face Accelerate using \textit{torch.compile} (gray, ``Baseline'') reached a maximum batch size of 16 on an NVIDIA 80GB A100 before Out of Memory (OOM) errors were encountered. The equivalent model trained with the BioNeMo Framework (green) reached a maximum batch size of 46 before OOM errors.}
    \label{esm2-batch-scale}
\end{figure}
\begin{figure}
    \centering
    \includegraphics[width=1\linewidth]{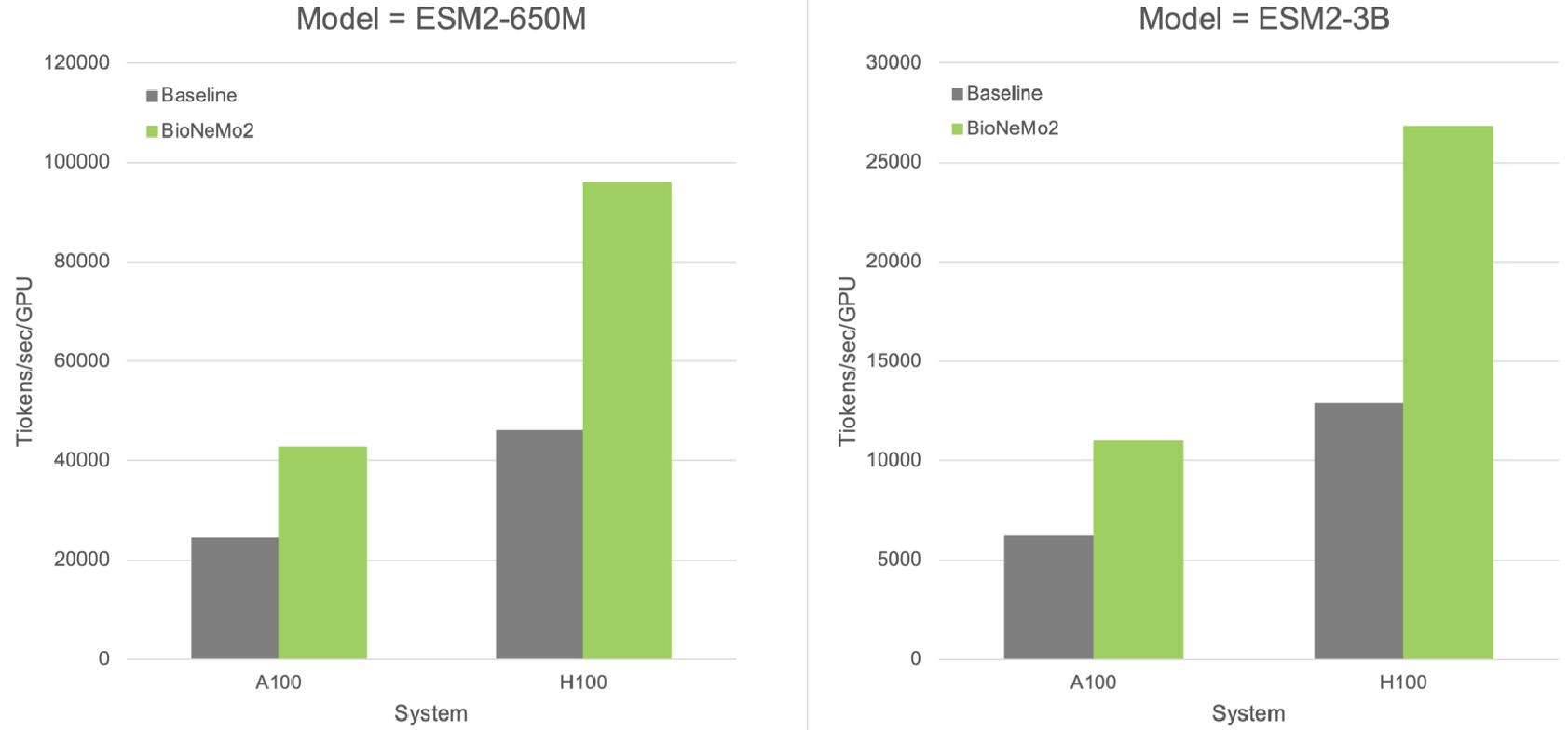}
    \caption{The BioNeMo Framework enables higher-throughput training across a range of device types and model sizes. The ESM-2 650M or 3B parameter model trained with Hugging Face Accelerate using \textit{torch.compile} (gray, "Baseline") is compared to the equivalent model trained with the BioNeMo Framework (green). Results are shown using the largest possible batch size for each model and framework and training on 16 NVIDIA A100s or H100s split over two compute nodes.}
    
    \label{esm2-device-scale}
\end{figure}

To assess the performance of the BioNeMo Framework, we compared training speeds for the ESM-2 model at various sizes and in different parallel configurations. First, we looked at single-device throughput for the 33-layer, 650M parameter variant. Benchmark speeds for the model as implemented in the Hugging Face Transformers library (4.44.0) were obtained using the Accelerate training library (0.33.0), using \textit{torch.compile’s} default dynamo backend, running inside the NVIDIA Deep Graph Library (DGL) PyTorch container\footnote{Access via \href{http://nvcr.io/nvidia/pytorch:24.07-py3}{nvcr.io/nvidia/pytorch:24.07-py3}}, PyTorch version 2.4.0). Training was performed on a single NVIDIA A100 GPU. Batches were increased until a CUDA out-of-memory (OOM) error occurred after a batch size of 16, with the resulting training throughput at each batch size shown in Figure \ref{esm2-batch-scale}. The BioNeMo Framework training throughputs were obtained on the same hardware with version 2.0 of the framework container. A maximum batch size of 46 was possible before an OOM error occurred, with an overall single-device training throughput 1.47x faster than that obtained with the Transformers library. BioNeMo Framework achieved a 59.2\% model flops utilization (MFU) \cite{Chowdhery2022-hg}, while the model trained with the Accelerate library reached an MFU of 40.1\%.

Next, we quantified how BioNeMo Framework performed in distributed training, at larger model sizes, and on H100 GPUs in addition to A100 GPUs. When training on 16 devices split across two compute nodes, the acceleration observed over the baseline training examples remained consistent. Training the 3B parameter variant of ESM-2 yields an MFU of 62.3\% for BioNeMo Framework on A100 GPUs, which drops to 49.1\% on the H100 system (a speedup of only 2.43x compared to the theoretical maximum of 3x). Hugging Face models were trained using DeepSpeed (0.15.1) with ZeRO level 1 using distributed data parallel \cite{Rajbhandari2019-wp}.

To determine how efficiently model training can be accelerated through parallelization, we looked at training speed for the 3B parameter model as we increased the number of GPUs. Each library's single-node (8 A100 GPUs) training throughput was used as a reference for perfect ‘linear’ scalability. Next, we measured training throughput at 2 nodes (16 GPUs), 4 nodes (32 GPUs), 8 nodes (64 GPUs), 16 nodes (128 GPUs), and finally 32 nodes (256 GPUs), as shown in Figure \ref{esm2-device-scale}. The BioNeMo Framework achieved a training speed at 256 GPUs that was 96.9\% of the extrapolated single-node throughput corresponding to a 60\% MFU and a total time to train over 1 trillion tokens of just 4.2 days. The models trained using DeepSpeed reached 80\% of their extrapolated single-node performance, with a total time to train over 1 trillion tokens of 8.8 days.

\subsection{BioNeMo Single-Cell Data Loader}
BioNeMo-SCDL provides an independent PyTorch-compatible dataset class for single-cell data with a consistent API. This package can be run independently from BioNeMo. It is designed to accelerate the training of transformer models with single-cell data. A data loader utilizing the Single-Cell Data Loader (SCDL) interface provides noticeable acceleration over comparable AnnData Loaders. SCDL does not require loading data into memory, however, it runs 1.1 to 2.75 times faster than a comparable AnnData loader. SCDL is also easy to customize and currently includes rank-order transformations for seamless integration into training Geneformer-like single-cell foundational models. It also supports pulling in user-defined metadata from the AnnData source files, which makes it highly adaptable for fine-tuning workflows. SCDL does all this by efficiently storing and accessing data from compressed sparse matrices, which are an integral part of single-cell datasets, using NumPy's memory-mapped format. This allows it to handle arbitrarily large datasets rapidly, unlike other single-cell data loading approaches. Currently, SCDL supports conversion from the AnnData format, but the conversion code is modular to facilitate the development of other custom converters in the future. BioNeMo SCDL's API resembles that of AnnData, making it is easy to use for practitioners familiar with the AnnData API.
\begin{figure}
    \centering
    \includegraphics[width=0.5\linewidth]{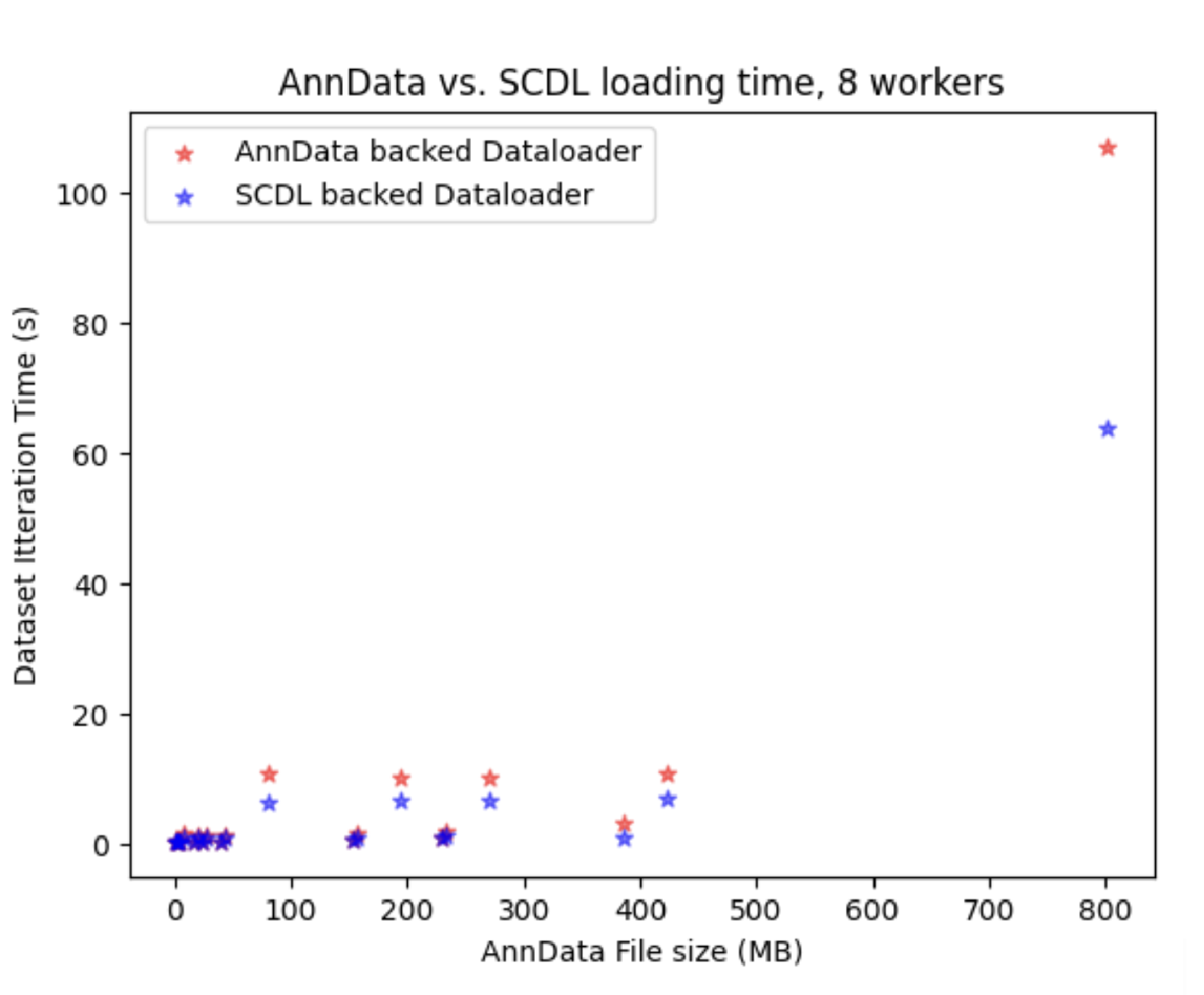}
    \caption{Dataset iteration time for a data loader using SCDL and an AnnData data loader. SCDL shows a consistent speed-up. }
    \label{fig:scdl}
\end{figure}

\subsection{WebDataModule}
The WebDataset is a popular Python library for high-performance data loading in training neural network models. It implements stream processing of input data with parallel data access via data sharding. It allows streaming data from local storage as well as cloud storage endpoints\footnote{https://github.com/webdataset/webdataset}. On the other hand, the \textit{LightningDatamodule} is the standard way of preparing, loading, and presenting data to the training workflow in the Lightning framework. A customized \textit{LightningDatamodule} requires user implementation of a few Lightning APIs, which could be cumbersome in the case of \textit{WebDataset}. In particular, WebDataset requires a specific file name scheme in the input data shards. It also requires a specific type of Python generator chained together to form the data transformation used in the dataset and data loader. To make it easy for the user to leverage WebDataset, we implemented the \textit{webdatamodule} in the BioNeMo Framework to create LightningDatamodule specifically from WebDataset. There are two components in our WebDataModule: 1) a general ``WebDataModule'' class. This is a subclass of the \textit{LightningDatamodule} and maps various WebDataset inputs, e.g., the data shard directory and the data file naming scheme, to the corresponding LightningDatamodule setup. 2) a ``PickledDataWDS'' that inherits from the WebDataModuleclass mentioned above with the additional built-in utilities to create the WebDataset data shards from Python pickle data objects. Here we refer to the BioNeMo Framework documentation for the webdatamodule usage and examples.

\subsection{Size-aware Batching}
Many applications of neural network models require processing data of varying dimensions. Especially in the case of GNNs, the input data exhibit variations in their graph topologies, which are defined by the structure of their node sets and edge sets. In drug discovery and molecular modeling, graphs can represent the molecular bond structure of proteins or small molecules and the general spatial interactions among the atoms or higher-level chemical structures \cite{David2020-gz}. Instead of processing the data samples individually, processing them in batches is often more efficient for better hardware utilization via parallel data processing and for more stable gradient estimation during training, which can reduce training times. Perhaps the most obvious way of batching graphs is via padding, where a multi-dimensional array represents certain graph properties, with the leading dimension being the batch dimension, i.e., indexing the samples in the batch, while the remaining dimensions represent the inherent feature dimensions of the graph property in question. Padding is typically required in this scheme so that slicing along the batch dimension results in feature arrays of the same size. At the cost of extra memory space -- and perhaps extra processing time due to the padding values -- the padded array representation is convenient in indexing or slicing the samples, and it is usually machine learning framework friendly where a large number of arithmetic operations have been developed to work on dense arrays. A more memory-efficient approach to batching graph data leverages a graph adjacency matrix’s sparse representation to collate multiple graphs into a disjoint meta-graph, where sparse arrays represent the meta-graph itself. This approach typically concatenates samples’ array of a graph property, e.g., node features, into a single array and can discern the operations applied on different graphs in the batch via auxiliary data like an array that masks the individual graphs in the concatenation. No padding is required, and efficient algorithms have been developed to parallelize the sparse batch of samples subgraph sampling and message-passing techniques, among others. This approach is widely used in GNN frameworks such as PyG \cite{Fey2019-hu} and DGL \cite{Wang2019-ch}. While this approach is memory efficient and often saves unnecessary computation compared to the padding approach, it can become unintuitive as to what utility functions can be used for certain operations and what the time and space complexities are for those operations as a function of the input data size.

In the BioNeMo Framework, we have developed tools to help address memory utilization issues often arising from batching data of varying dimensions. Specifically for the application of GNN model training, we have developed the \textit{size-aware batching} Python module to make GPU memory utilization more efficient. There are two main components in the size-aware batching module: a size-aware batcher and a bucket batch sampler.

\subsubsection{Size-aware batcher}
As mentioned previously, mini-batching is essential in training neural networks. In a static batch size scheme, where the same number of samples is batched on every training step, the mini-batch size is often deemed a critical hyper-parameter to be tuned, considering the tradeoff between memory consumption and training efficiency. In the case of GNN training using the sparse mini-batch approach, it is often challenging to compute the relationship between batch size and the resulting memory consumption due to the inherent heterogeneous computational requirements among the samples. Too small a batch size can lead to volatile gradient estimation that can lower the training efficiency, and too big a batch size can often lead to aborting the entire batch due to out-of-memory error. To address this problem, we have developed the size-aware batcher module to adjust the batch size in real-time during training to maintain a relatively constant level of GPU memory consumption. There are three critical steps in the size-aware batcher: (1) a utility function called \textit{``collect\_cuda\_peak\_alloc''}that collects CUDA memory allocation for a given workflow and can optionally output some features the user may want to use in fitting or profiling the CUDA memory usage; (2) a user implementation of memory allocation model that can be trained to predict memory consumption of the workflow based on the features as described in the previous steps; and (3) A Python Iterable that loops over an input sequence of data samples and outputs mini-batches of samples, ensuring none of the batches exceed CUDA memory capacity by accumulating the memory consumption requirements from each sample based on the memory predictor from step (2). While the Python Iterable in step (3) is designed to work directly in PyTorch iterable-style dataset, a size-aware batch sampler is also implemented to make it easy to adapt the size-aware batching method in PyTorch’s indexable or map-style dataset.

\subsubsection{Bucket batch sampler}
In the padding-based mini-batching approach, minimizing the amount of padding in creating the data property arrays is often desirable. Under the assumption that sufficient data samples exist in each subset of data size,  the padding minimization problem can be solved by mini-batching data of similar sizes. We have implemented the bucket batch sampler, a specialization of the PyTorch \textit{BatchSampler}, that bins the data samples into buckets of desirable sizes and returns the resulting indices to the data sample in each bucket. The size measurement of each data sample and the lower and upper boundaries of each bucket are taken from user input. We have also provided a utility function \textit{``create\_buckets''} to help build buckets given input constraints, such as their maximal width and the minimal number of samples each can contain. Smaller bucket width translates into more homogeneous data sizes in the resulting mini-batches, while the minimal sample count for each bucket is a fuzzy way to control the bucket size.

\subsubsection{Benchmarks and results}
\begin{figure}
    \centering
    \includegraphics[width=0.5\linewidth]{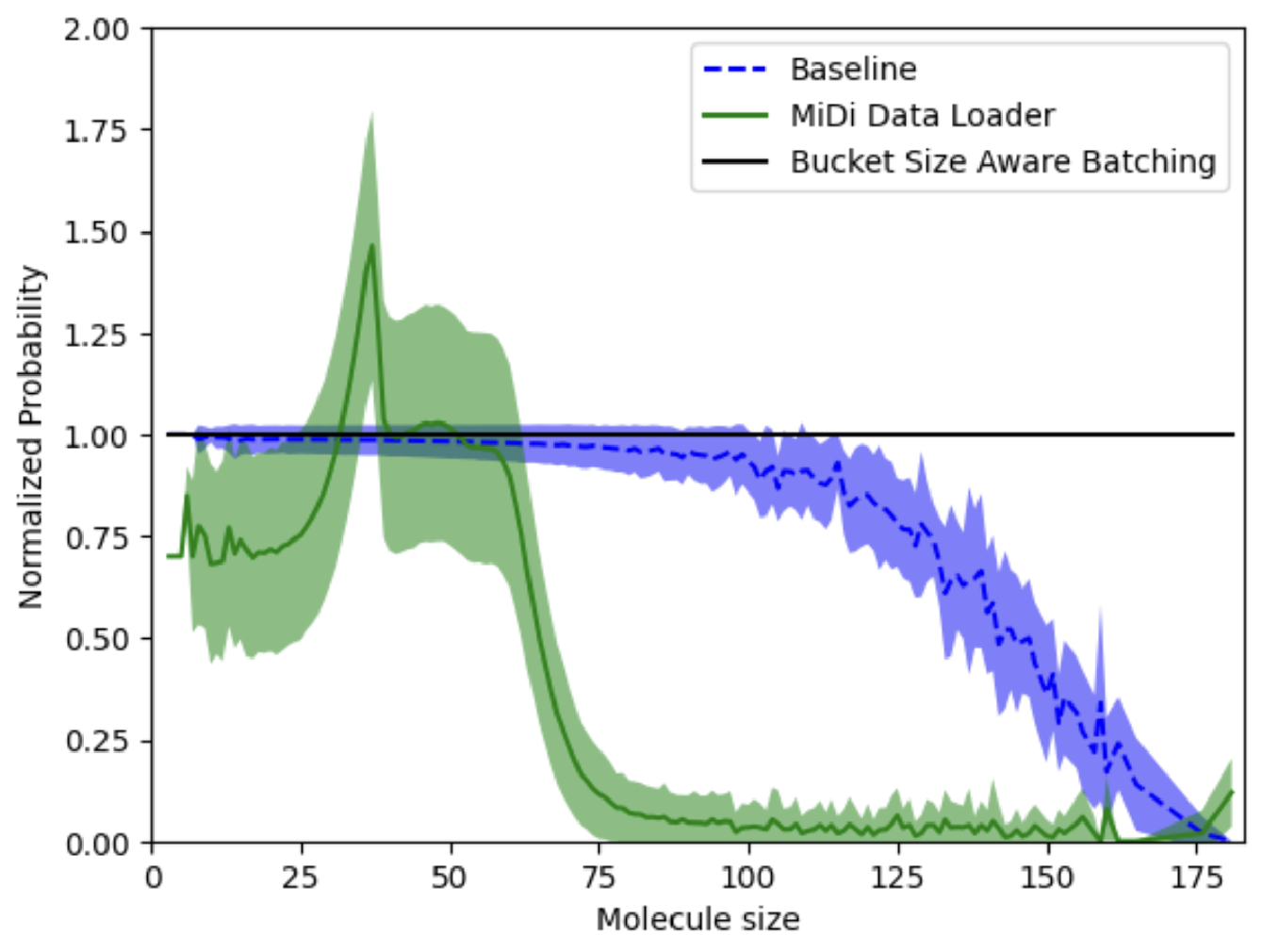}
    \caption{Shown here is the probability of samples of different sizes in the GEOM-Drug dataset. The MiDi and the baseline static size batching often undersample the moderate to large data for different reasons. The large samples are skipped in the baseline method due to OOM error. In the MiDi method, sampling is biased towards the most prominent sizes by the design of the method. The bucket size-aware batching gives a uniform distribution across a wide spectrum of data sizes, i.e., it does not introduce sampling bias.}
    
    \label{fig:mol-size}
\end{figure}

Here we demonstrate the efficiency of size-aware batching in the GEOM-Drug data \cite{Axelrod2022-lo} by comparing three different data loading methods: the baseline data loading with static batch size, the MiDi data loader used by the Megalodon model, which implements an adaptive batch size \cite{megaloden2024}, and our data loader using the size-aware batcher that takes the input samples from the buckets of the bucket batch sampler. We generate as many batches as 10 times the dataset size to simulate 10 training epochs using each method. We then examine these batches of data in terms of their data size or molecular size as well as the amount of padding involved. \ref{fig:mol-size} shows the probability of samples of different sizes in the GEOM-Drug dataset. The MiDi and the baseline static size batching often undersample the moderate to large data for different reasons. The large samples are skipped in the baseline method due to OOM error. In the MiDi method, sampling is biased towards the most prominent sizes by the design of the method. The bucket size-aware batching gives a uniform distribution across a broad spectrum of data sizes, i.e., it does not introduce sampling bias.

\begin{figure}
    \centering
    \includegraphics[width=0.5\linewidth]{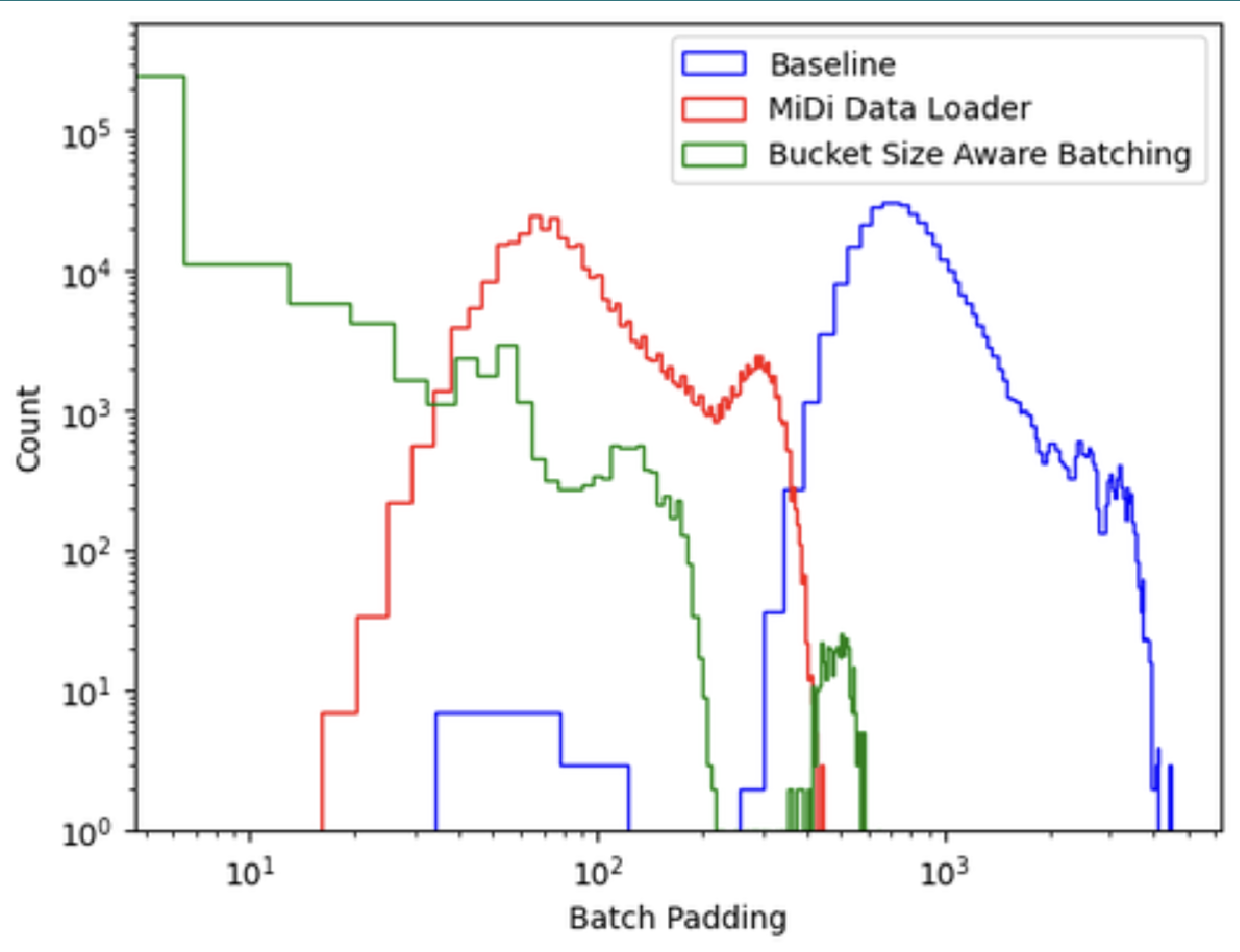}
    \caption{A histogram of the numerical elements required for padding the batches. The bucket size-aware batching requires very little padding in most batches while the MiDi and baseline methods require moderate and large amounts of padding respectively.}

    \label{fig:batch-pad}
\end{figure}

We then compare the amount of memory padding required to generate data batches from the three methods in question. In \ref{fig:batch-pad}, we create a histogram with the numerical elements required for padding the batches. The bucket size-aware batching requires very little padding in most batches, while the MiDi and baseline methods require moderate and large amounts of padding, respectively.

\section{Discussion}
\subsection{Contributions from the Research Community}
As open-source software, the BioNeMo Framework is a community-driven platform that enables users to contribute across various biomolecular domains and optimizes performance on GPU hardware. There have already been many contributions, from protein design workflows to DNA data loading, derived from real-world applications in drug discovery.

Dyno Therapeutics, a biotechnology company applying AI to discover novel gene therapy vectors with transformative delivery properties, integrated parameter-efficient fine-tuning to the BioNeMo Framework's ESM-2 implementation by enabling freezing of specific layers and only fine-tuning the most relevant parts of the network. Traditional fine-tuning requires updating the entire model, which can be costly and inefficient. In contrast, this approach reduces the risk of overfitting and significantly speeds up training, making it more resource-efficient.

Flagship Pioneering is harnessing AI to drive greater innovation in the life sciences through their initiative Flagship Intelligence, which added protein inverse folding model RL-DIF\cite{ektefaie2024reinforcementlearningstructureconditionedcategorical} to the BioNeMo Framework. Unlike traditional protein folding models, which predict a structure from a given sequence, RL-DIF does the reverse: it generates diverse sequences that can fold into a specific target structure, allowing researchers to design new proteins that fit precise structural requirements and opening vast possibilities in drug discovery and biotechnology.

Relation Therapeutics leverages AI with a Lab-in-the-Loop approach to deepen our understanding of disease biology and therapeutic discovery and has added a new DNADL data loading tool to the BioNeMo Framework for simplifying data loading of common genomics data formats. This allows users to easily develop AI models for DNA sequences, including enabling the training of models on specific genome regions or sequences with known variants for more precise insights.

Finally, Weights \& Biases, the leading AI developer platform for experiment management when training and fine-tuning models, has also integrated their tooling into BioNeMo Framework to automatically enable experiment management for users, tracking training runs across restarts and help users monitor their experiments more effectively.

\subsection{Scaling BioNeMo Framework Training with Cloud Computing Resources}
The scalability of cloud services allows BioNeMo Framework to support a wide range of project requirements and large datasets without the need for extensive on-site infrastructure. This environment enables teams to access and customize pre-trained models, making it feasible to conduct proof-of-concept studies before deployment to a production environment. For example, BioNeMo Framework is fully compatible with Amazon Web Services (AWS) infrastructure, providing scalable compute resources, including EC2 instances with NVIDIA GPUs, to meet the computational demands of training. AWS storage solutions, such as S3, enable data access and management throughout the training process. This cloud-based setup supports iterative model refinement, eliminating the need for on-premise hardware and accelerating the development of models tailored to specific research objectives in biopharma.

A-Alpha Bio, a biotechnology company specializing in protein-protein interaction (PPI) prediction, has used the BioNeMo Framework on AWS to enhance its computational pipeline for drug discovery. By deploying BioNeMo on Amazon EC2 P5 instances equipped with NVIDIA H100 Tensor Core GPUs, A-Alpha Bio achieved a 12-fold increase in inference speed and processed over 108 million inference calls in two months — significantly higher than their initial target. The increase in inference capacity enabled the evaluation of 10 times more protein binding predictions, thereby expanding the scope of potential drug candidates under consideration. Integrating BioNeMo with AWS Batch allowed A-Alpha Bio to scale their computational workflows efficiently. By using AWS's capacity reservation options, such as EC2 Capacity Blocks, they secured predictable access to high-performance compute resources. This setup enabled iterative model fine-tuning and optimization, reducing the need for multiple experimental cycles, which lowered overall costs and accelerated the process of therapeutic design.

Using the BioNeMo Framework also facilitated A-Alpha Bio's exploration of a wider mutational landscape, improving the likelihood of identifying high-affinity binders for therapeutic applications. This approach reduced the number of wet-lab experiments required in the design-build-test cycle, allowing them to shift more of their exploration to computational testing. Consequently, they were able to explore more complex protein modifications, statistically increasing their chances of discovering potent therapeutic candidates.

As part of this effort, A-Alpha Bio also contributed their notebook for zero-shot protein design with ESM-2 back to the BioNeMo Framework codebase, now available for other users to deploy. This leverages masked language modeling to predict likely amino acid substitutions at specific positions, empowering researchers to efficiently design and optimize proteins for anyone aiming to perform fine protein perturbations for downstream applications.

\section{Conclusion}
The BioNeMo Framework is a significant step in enabling drug discovery workflows through a robust, open-source ecosystem of GPU-accelerated tools. With its demonstrated capability in training throughput and scalability, coupled with its modular and extensible design, BioNeMo Framework has the potential to enable further advancements across diverse biomolecular domains. In the future, the BioNeMo Framework team plans to continue refining the framework APIs to facilitate modularity, component-level reusability, customization, and experimentation. We aim to support our users in being able to contribute framework-level improvements back to the project while independently sharing and maintaining packages that make use of the framework. We see this as an evolving collection of sub-packages that may grow or shrink as improved abstractions become available. We aim to prioritize workflows that demonstrate our framework’s strengths, namely the ability to train, infer, and fine-tune with very large models that do not fit in single devices. We will focus future efforts on pushing the envelope with state-of-the-art performance in these settings. As BioNeMo uses other actively used frameworks within NVIDIA, including NeMo, Megatron, and Transformer Engine, it will continue to benefit from performance and usability improvements across the NVIDIA software ecosystem.

\printbibliography

\end{document}